# Parsimonious Inference on Convolutional Neural Networks: Learning and applying on-line kernel activation rules

*I. Theodorakopoulos, V. Pothos, D. Kastaniotis and N. Fragoulis[1], Irida Labs S.A,  January 2017*

**Abstract**

*A new, radical CNN design approach is presented in this paper, considering the reduction of the total computational load during inference. This is achieved by a new holistic intervention on both the CNN architecture and the training procedure, which targets to the parsimonious inference by learning to exploit or remove the redundant capacity of a CNN architecture. This is accomplished, by the introduction of a new structural element that can be inserted as an add-on to any contemporary CNN architecture, whilst preserving or even improving its recognition accuracy. Our approach formulates a systematic and data-driven method for developing CNNs that are trained to eventually change size and form on-the-fly during inference, targeting to the smaller possible computational footprint. Results are provided for the optimal implementation on a few modern, high-end mobile computing platforms indicating a significant speed-up of up to x3 times.*

1. **Introduction**

Deep learning [**BLH**] has emerged as the dominant approach for performing various classification tasks ranging from computer vision to speech processing **[Krizhevsky]**, **[Graves]**. Relying on deep convolutional neural networks architectures, their success is due to their ability to learn abstract and high-level feature representations using large amounts of training data in an end-to-end fashion. However, top-performing systems usually involve deep and wide architectures and therefore come with the cost of increased storage and computational requirements, while the trend is to continuously increase the depth of the networks.

Besides these drawback, deep neural networks have a very intriguing property, which provides opportunities for optimizations regarding both storage space and computations-the redundancy of parameters and computational units (e.g. convolutional kernels). **[Cheng], [Ba], [Han2015].** This redundancy, often leads to a misfit between the machine-learning task indented to be carried out and the CNN structure selected for this task.

At the same time, there is an increasing need to use deep CNNs in applications running on embedded devices. These devices-especially in the Internet of Things (IoT) era- are usually equipped with small storage and low computational capabilities. As such they cannot neither store large CNN models nor to cope with the associated computational complexity of a deep and wide CNN. Therefore, it becomes apparent that, in order to make CNNs appealing for mobile devices, both the parameter size and the number of kernels need to be reduced. But then, a fundamental dilemma becomes apparent:

---
[1] Corresponding author, email: nfrag@iridalabs.gr

> ***Dilemma:*** *Can we design computationally efficient CNNs and at the same time to maintain recognition accuracy and generalization, while keep the overall structure efficient for operating on mobile embedded devices?*

This work tries to answer the above-mentioned dilemma, by providing a systematic way for implementing CNN variants that are **parsimonious** in computations. To this end, the proposed approach allows training a CNN to:

i) Use as few computing resources as possible. The devised procedure results into an **optimal pruning of a CNN architecture,** guided by the complexity of the task and the nature of the input data.
ii) **Change size and form on-the-fly during inference,** depending on the input data. This property enables to perform inference using less effort for "easier" instances of data than others.
iii) Optimize for the above objectives via **regular back-propagation**, **simultaneously** to the primary task objective of the model. This way we avoid the prune-fine-tune iterative procedure, which is usually followed in order to reduce the size of a model.

The proposed framework incorporates a new learning module, the **Learning Kernel Activation Module (LKAM)**, serving a dual purpose: Enforcing the utilization of less convolutional kernels by learning kernel activation rules and de-activate a sub-set of filtering kernels during the inference phase, depending on the input image content and the learned activation rules. Using this module, the CNN essentially learns how to reduce its initial size on-the-fly, by exploiting the presence of multiple information flow paths, thus maximizing computing economy during inference. Since a reduction in the number of applied kernels in any layer leads to the reduction of channels passed into the next layer, the reduction of the overall computational load is even more important.

The proposed method is compatible with any contemporary deep CNN architecture and can be used in combination with other model thinning approaches (optimal filtering, factorization, etc.) resulting into a further processing optimization.

In the following **Section 2,** we present some related works and in **Section 3** we present the proposed method. In **Section 4** we present some experimental results based on simulations and on **Section 5** speed results obtained through porting to some modern high-end mobile platforms. In **Section 6**, some basic ideas are discussed for the VLSI implementation case.

## 2. Related Work

Deep learning primarily developed as a tool to find meaningful representations from large collections of data. In order to achieve this, a complex function of the data is learnt using a large sequence of simple functions, which in turn results in a large number of parameters. These simple functions however are both computational and memory intensive. Therefore, this initial approach contradicts modern applications where power consumption and inference time play a major role. In particular, for the case of IoT applications the overall computational load as well as the total number of memory transactions might become prohibitive.

To this end, the reduction of the computational load associated with a specific deep-learning structure is the enabling factor towards the broadening of the application field of these structures to IoT and in general in applications featuring a system with low computational capabilities.

In this direction, most of the researchers attempt to exploit the data sparsity and the redundancy of the parameters inherent in CNNs in order to prune some parts of the convolutional network and thus ease the computational load of the overall structure, in an off-line, post-training approach. In some methods, the coefficients of a CNN are analyzed after training and some of them are zeroed according to their magnitude- leading to sparse matrices exploitable by sparse arithmetic software. In some others, the CNN is trained in such a way so to result on a set of coefficients containing as many insignificant coefficients as possible.

In a data-driven approach, **[Hu16]** proposed a method which iteratively optimizes the network by pruning unimportant neurons based on analysis of their outputs on a large dataset.

Feng et al. **[Feng15]** proposed a method for estimating the structure of the model by utilizing un-labelled data. Their method called Indian Buffet Process CNN (ibpCNN), captures the distribution of the data and accordingly balances the model between complexity and fidelity.

Similarly, Wen et al. **[Wen16]** incorporated Structured Sparsity Learning (SSL) in order to regularize the number of filters (and their shapes), the number of channels and the depth of the network. From an implementation perspective, SSL also targets to the formulation of a dense weight matrix in order to completely remove channels, filters or even whole layers.

Yang et al. **[Yang]** proposed an energy-aware pruning algorithm for CNNs that directly uses energy consumption estimation of a CNN to guide the pruning process. For each layer, the weights are first pruned and then locally fine-tuned with a closed-form least-square solution to quickly restore the accuracy.

Authors in [**Han2015**] proposed a three-step method, which allowed them to prune redundant connections without affecting the accuracy. In the first step, they train a network to learn which connections are important. In the second stage, connections characterized as unimportant are pruned and in the last stage, the network is re-trained in order to fine-tune the weights.

Similarly, in **[PeforatedCNNs]** authors targeting to implementations for low power devices, by taking advantage of the sparsity immanent in intermediate filter responses in order to reduce the spatial convolution at every layer. More specifically, they are inspired by the loop perforation technique (originally proposed for source code optimization) in order to skip the convolution operation at several locations.

### 3. Parsimonious inference

The target of the framework proposed in this paper is to implement a CNN structure able to learn its primary task, while being economical on both size and complexity. Since the main source of computational load in a CNN is the number of convolutional kernels employed in each convolutional layer, the idea exploited in this work is to enforce channel-wise sparsity to the outputs of each convolutional layer. That way, each kernel either learns how to capture useful information or vanishes.

In contrast to the regularization approaches [**Wen16**], which try to enforce a global sparsity pattern in order to prune kernels and channels with zero output values, we propose a technique that enables kernels to learn information which may be useful to a subset of the observed cases. One step forward, by enforcing this sparsity via simultaneously learned, data-driven kernel activation rules, the same rules can be used during inference in order to avoid computing kernels which are not useful for a particular datum. That way, only the relevant kernels are computed, resulting to a significant economy in processing time and power. At the end of the training procedure, kernels that have not managed to learn features that are relevant to any of the data, resulting to zero utilization, can be permanently pruned from the model.

*3.1. The learning kernel-activation module.*

The idea is presented in Figure 1. In this figure, a part of a typical convolutional network is shown, depicting only the i-th and the (i+1)-th convolutional layer.

In this figure, a Learning Kernel-Activation Module (LKAM) is introduced, linking two consecutive convolutional layers. This learning switch module is capable to switch on and off individual kernels of any layer depending on its input, which is the output of the previous convolutional layer.

The module learns which kernel to disable during the CNN training process, which is for that reason specifically devised to facilitate such operation by exploiting data sparsity usually employed in images.

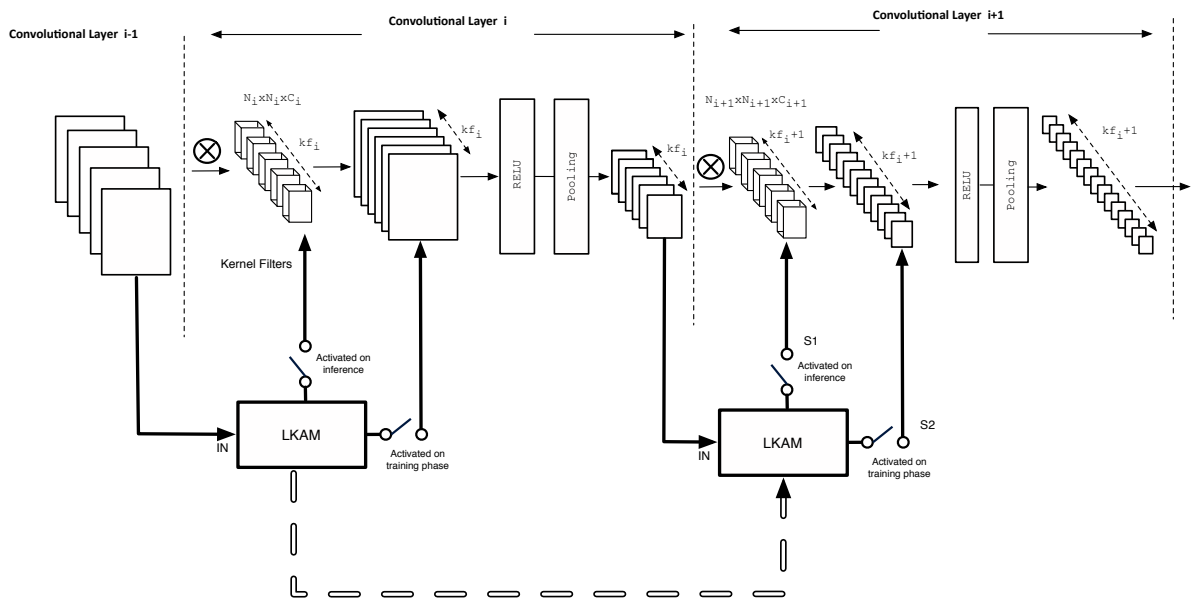

Figure 1: The Learning Kernel-Activation Module (LKAM) is introduced between successive convolutional layers of a CNN.

The aim of these modules is initially to induce the desired channel-wise sparsity to the feature maps. Simultaneously, LKAM learns an activation rule for each kernel, which is later been exploited during the inference phase. Many types of activation rules can be formulated using regular differentiable functions, available in all deep-learning frameworks. In this work we study one of the simplest and lightweight rules, constituting by a bank of 1x1 convolutional kernels followed by average pooling and a sigmoid function that that offers a smooth and differentiable transition between active and inactive state. The choice of this rule was made in order to keep computational overhead of the LKAM modules as low as possible. The internal structure of the LKAM module is shown in Figure 2.

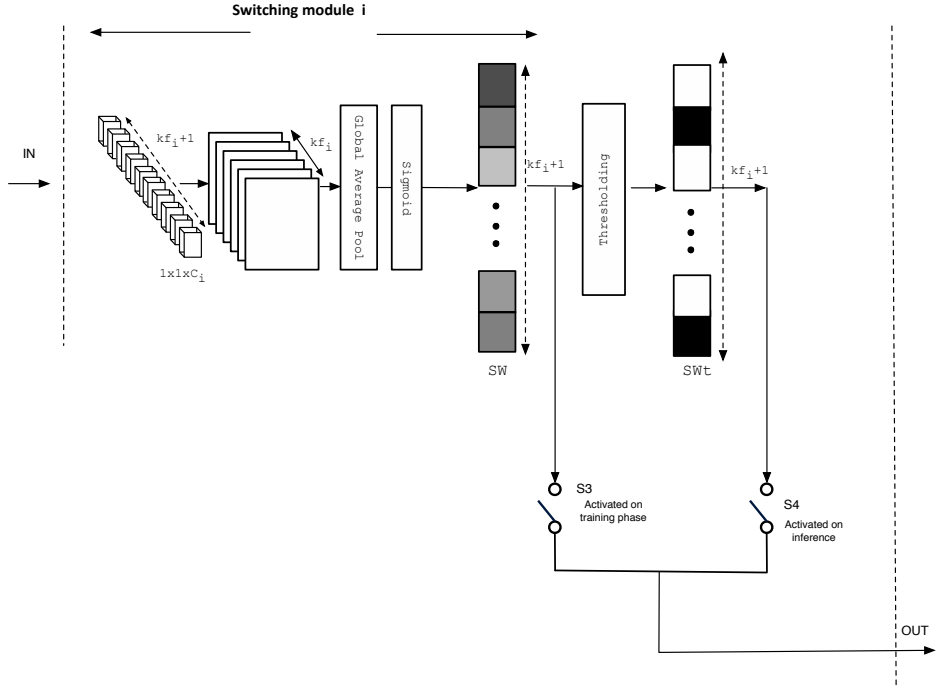

Figure 2: The internal structure of a LKAM module

First, the feature maps of the i-th convolutional layer are fed into this module. These are processed by $kf_{i+1}$ kernels of size $1x1xC_{i+1}$ resulting into $kf_{i+1}$ feature maps. These maps are then fed into a global average-pooling block, which averages the values of each feature map producing a single number for each feature map. Each one of this numbers is then passed into a sigmoid layer implementing the function

$$f(x) = \frac{1}{1+e^{-k(x-x_0)}} \qquad (3)$$

In this way a vector $SW = \{sw_1, sw_2, \dots, sw_{k_{fi+1}}\} \in \mathbb{R}^{k_{fi}+1}$ with values between 0 and 1, is formed.

The elements of this vector are used in the training phase, through the switch S3, in order to multiply the values of the corresponding feature map in the (i+1)-th convolutional layer, thus imposing the desired sparsity. During this phase, switches S2 in Figure 1 and S3 Figure 2 are activated, while switches S1 Figure 1 and S4 in Figure 2 are deactivated. This way, the information flow is tweaked by enforcing certain feature maps to gradually have smaller influence to the overall network under the corresponding rules, which are in turn co-adapting. The goal of the training process is to obtain the combination of kernels and activation rules that produce the sparsest $SW$ vectors possible. The learned rules can indicate the kernels with zero influence so that the corresponding kernels to be excluded from computation.

*3.2 Training procedure*

The training of the LKAM modules takes place concurrently to the training of the rest of the network by using back-propagation and aims to the calculation of the coefficients for the $kf_i$, 1x1 convolutional masks in the LKA modules.

In order to impose the desirable channel-wise sparsity, the primary loss function used during back-propagation has to be augmented with a new term, penalizing the use of convolutional kernels. The easiest way to achieve this is to add a term proportional to the L1 norm of the SW vectors, denoted as $L_{aug}$ and given by the following equation:

$$L_{aug} = \frac{G_i}{2m}\sum_i |sw_i| \qquad (5)$$

where $G_i$ is a gain factor and $m$ is the length of the vector. The overall loss now becomes:

$$L(w,b,sw) = L_t(w,b) + L_{aug}(sw) \qquad (6)$$

Where $L_t(w,b)$ is the main loss, dictated by the primary task of the model (e.g. Hinge loss, Euclidean etc.). By tuning the gain factors $G_i$ we can control the sparsity level separately for each layer.

*3.3. Pruning kernels*

Past the end of the training phase, a simple statistical analysis of the kernel activations throughout the dataset can reveal kernels with very low contribution, characterized by zero or near-zero utilization. Such kernels can safely be considered redundant and pruned from the model along with their corresponding filters in the LKA modules.

*3.4 Real-time deactivation of kernels during Inference (recognition) phase*

During *inference*, the elements of the vector SW are used as a set of switches that control the corresponding kernels in the (i+1)-th convolutional layer, depending on the input from the i-th layer (Figure 5). Since the value of each $sw_i$ can be any real number between 0 and 1, a simple thresholding is used as the activation criterion, where the elements of the vector SW are binarized (i.e. forced to take values 1 or 0) using a threshold value **thres** as follows:

$$sw_i = \begin{cases} 0, & sw_i < thres \\ 1, & sw_i \geq thres \end{cases} \qquad (4)$$

The resulting binary activation values are the indicators of whether to apply the corresponding filtering kernels on the input data or skip the particular computations. Note that during inference, switches S2 in Figure 1 and S3 in Figure 2 are considered as active while switches S1 Figure 1 and S4 in Figure 2 are considered inactive.

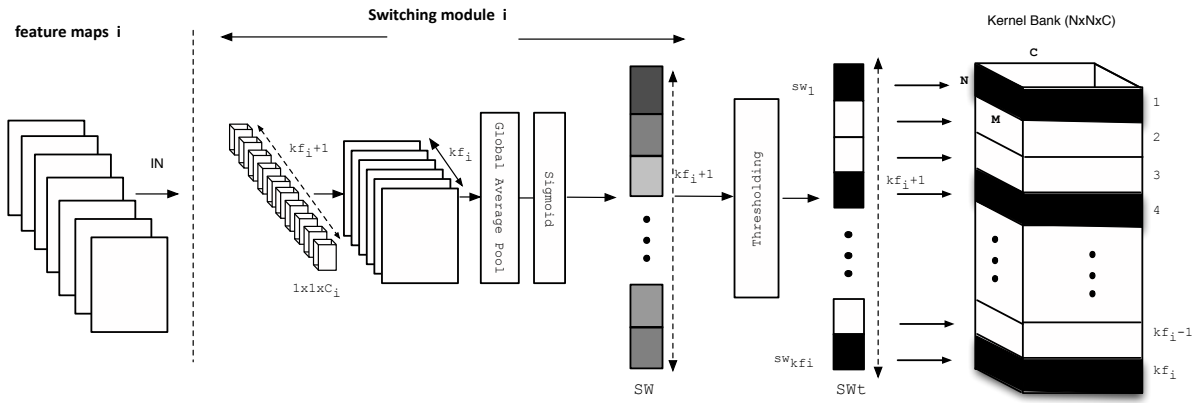

Figure 3: The switching process of the LKA module during inference

*3.5 Pruning/Deactivating Layers*

The computational gain achieved by kernels deactivation can be extended to layer-level, by using a residual architecture. In residual CNNs **[He]**, each convolutional layer is only responsible for, in effect, fine-tuning the output from a previous layer by just adding a learned "residual" to the input.

As illustrated in Figure 4, residual CNN architectures utilize layer bypass connections, which offer an alternative path for information flow between consecutive convolution layers. Such connections enable a complete deactivation of a convolutional layer without intercepting the subsequent processing stages.

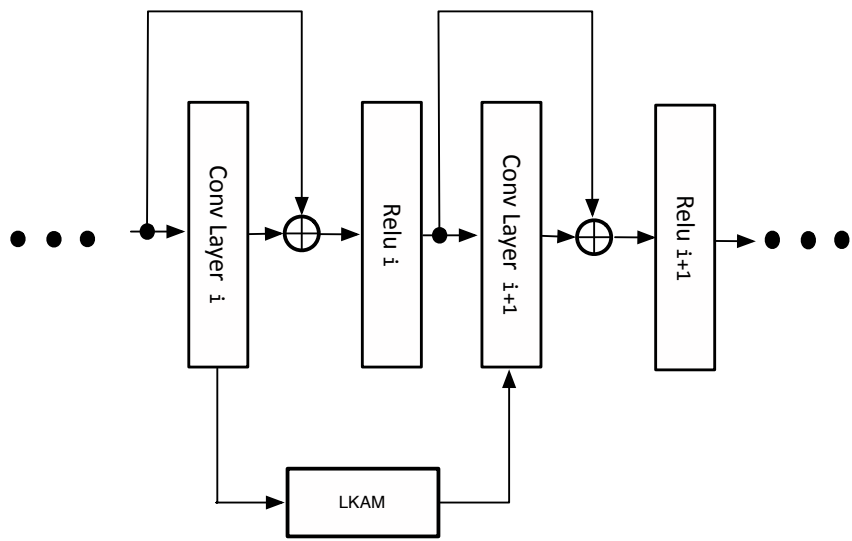

Figure 4: Application of the learning switch module technique in residual CNNs

## 4. Implementation and simulation

Our analysis has been conducted in two different classification problems: A food recognition problem, utilizing the FOOD 101 database **[Food101]** comprised by images of food organized to 101 categories and a general image recognition problem utilizing the ImageNet ILSVRC 2012 **[ILSVRC]** dataset comprised by images organized into 1000 categories. These two datasets have been chosen as two classification cases featuring different qualities. In particular, FOOD 101 problem is considered as a less complex classification case due to the smaller number of classes compared to the ILSVRC 2012 dataset, but featuring more abstract visual attributes since food styling can be very diverge. On the other hand, ILSVRC 2012 dataset contains images mainly depicting a single structured object, but similar classes are often discriminated by very fine details.

In the same spirit, two popular CNN architectures have been evaluated: CaffeNet and SqueezeNet 1.1 **[Squeeze]**. CaffeNet is almost identical to AlexNet [**AlexNet**] architecture, being a vanilla architecture with a medium-sized parameter space but relatively shallow. On the other hand, SqueezeNet is a deeper architecture designed for storage efficiency, consisting of 50 times less parameters compared to AlexNet, but exhibiting similar classification performance. The study of these two architectures that exhibit vastly different characteristics in terms of redundancy, is used to highlight the ability of the prosed framework to achieve the required computational parsimony under any circumstances.

In the case of CaffeNet architecture, 4 LKAMs have been used to control the activity of kernels in layers 2 to 5. Each of the LKAMs shares the same input as the layer that controls, in a configuration equivalent to the one illustrated in Figure 3. The implementation details of this structure are given in Table 1.

Table 1: Filtering and LKAM kernel sizes for the CaffeNet

| Layer | Kernel Size | LKAM Kernel Size | Activation Vector Size |
|---|---|---|---|
| conv1 | 11x11x96 | - | - |
| conv2 | 5x5x256 | 1x1x256 | 1x256 |
| conv3 | 3x3x384 | 1x1x384 | 1x384 |
| conv4 | 3x3x384 | 1x1x384 | 1x384 |
| conv5 | 3x3x256 | 1x1x256 | 1x256 |

For the SqueezeNet architecture, we use LKAMs for controlling the activity of the larger kernels that constitute the "Expand3x3" sub-modules inside the "Fire" modules 2 to 9. The LKAMs share the same input with the corresponding Fire modules they control, according to the configuration shown in Figure 5.

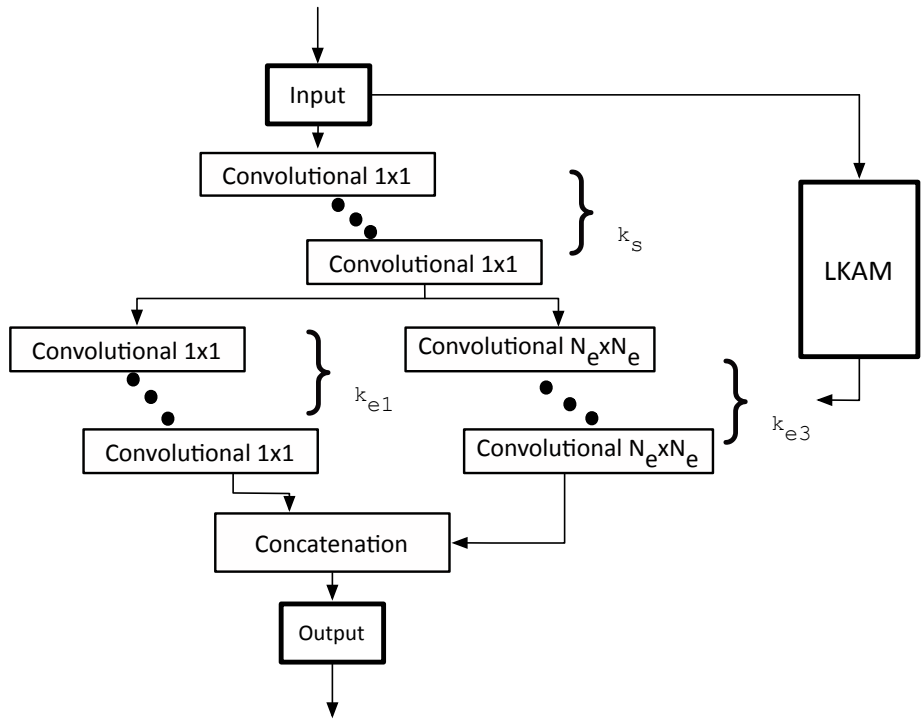

Fig 5: Modification of a Fire Module with the addition of a LKAM module: The target is to be able to deactivate computationally-intensive kernels aiming at the highest possible computational gain, so the obvious choice is to use the $N_e \times N_e$ filter bank (3x3 for the typical SqueezeNet). However, the 1x1 filter bank can also affected on specific occasions.

This configuration ensures the maximum possible gain from a potential deactivation of kernels, since a much more significant load corresponds to the larger kernels ($N_e$ is equal to 3) than the 1x1 kernels also present within the fire module. The detailed configuration for the SqueezeNet 1.1 architecture is summarized in Table 2.

Table 2: Filtering and LKA kernel sizes inside the relevant fire modules of SqueezeNet 1.1

| Layers | Kernel Size | LKAM Kernel Size | Activation Vector Size |
|---|---|---|---|
| conv1 | 3x3x64 | - | - |
| fire2/expand3x3 | 3x3x64 | 1x1x64 | 1x64 |
| fire3/expand3x3 | 3x3x64 | 1x1x64 | 1x64 |
| fire4/expand3x3 | 3x3x128 | 1x1x128 | 1x128 |
| fire5/expand3x3 | 3x3x128 | 1x1x128 | 1x128 |
| fire6/expand3x3 | 3x3x192 | 1x1x192 | 1x192 |
| fire7/expand3x3 | 3x3x192 | 1x1x192 | 1x192 |
| fire8/expand3x3 | 3x3x256 | 1x1x256 | 1x256 |
| fire9/expand3x3 | 3x3x256 | 1x1x256 | 1x256 |

All the experiments were conducted using the Caffe [Caffe] framework, on a server with dual Xeon CPU and two NVIDIA GTX 1070 GPUs. In order to compute the recognition accuracy during inference, kernel deactivation was emulated within the Caffe framework by performing a multiplication of the elements of SW with 10e4, prior to the application of the sigmoid function, in order to ensure that the final values of the SW vectors are either zeros or ones (simulating a threshold of 0.5). Subsequently those values were used to multiply the corresponding channels of the feature maps, using the same configuration with the training phase.

*4.1. Recognition Accuracy*

The recognition accuracy obtained by training the two architectures under the proposed framework is summarized in Tables 3 and 4. The threshold for the activation of kernels is 0.5 and the accuracy was measured on the validation set for ILSVRC and the test set for Food-101. The accuracy on the ILSVRC is compared to the reference models for CaffeNet[2] and SqueezeNet1.1[3] available on-line.

Table 3: Recognition accuracy for the CaffeNet and SqueezeNet on ILSVRC 2012 dataset

| Recognition Accuracy (%) | ILSVRC 2012 | | | |
|---|---|---|---|---|
| | Top1 | Diff. | Top 5 | Diff. |
| AlexNet Normal | 57.27 | - | 80.62 | - |
| AlexNet Switching | 58.46 | +1.19 | 81.21 | +0.59 |
| SqueezeNet 1,1 Normal | 57.59 | - | 80.44 | - |
| SqueezeNet 1.1 Switching | 59.59 | +2.0 | 82.05 | +1.61 |

Table 4: Recognition accuracy for the CaffeNet and SqueezeNet on FOOD 101 dataset

| Recognition Accuracy (%) | FOOD 101 | | | |
|---|---|---|---|---|
| | Top1 | Diff. | Top 5 | Diff. |
| AlexNet Normal | 68.54 | | 88.44 | - |
| AlexNet Switching | 68.86 | +0.32 | 88.61 | +0.17 |
| SqueezeNet 1,1 Normal | 65.65 | | 86.87 | - |
| SqueezeNet 1.1 Switching | 67.19 | +1.54 | 88.68 | +1.85 |

As it is evident, the introduction of an objective towards computational economy has not degraded the obtained accuracy on neither of the tester architectures. On the contrary, we observe a notable improvement of the classification accuracy compared to the reference models, on both datasets and both architectures. This reveals the dynamic of the approach regarding the overall control of the functionality of these CNNs.

*4.2. Computational Load*

The most important aspect of the presented framework though, is the improvement on the required computational load during inference. A detailed analysis of these two configurations is presented below.

---

[2] https://github.com/BVLC/caffe/tree/master/models/bvlc_reference_caffenet
[3] https://github.com/DeepScale/SqueezeNet/tree/master/SqueezeNet_v1.1

***Evaluation on CaffeNet*** has been contacted for both ILSVRC 2012 and FOOD 101 datasets. Engagement of the LKAM modules results on the reduction of the kernel filtering operations. This is shown graphically in the Figure 6 where the vertical axis corresponds to the activation frequency of a particular kernel throughout the validation set, while the horizontal axis corresponds to the kernels of each layer, sorted by ascending utilization (from left to right). For visualization purposes, the horizontal range is normalized and equal for all layers, even they accommodate different population of kernels. In such an illustration, a step-like plot implies kernels that are mostly permanently switched either off or on, while a smooth curve indicates kernels whose operation is data-dependent.

More specifically, the number of kernels that are calculated for each test image is significantly lower than the nominal number of kernels in the original networks. This dramatically reduces the related number of mathematical operations. It has to be noted again that the reduction on the active kernels in a single layer, besides the obvious benefit of avoiding the corresponding filtering computations, results into a respective reduction in the number of input channels into the next layer. This in turn, offers an additional computational gain, directly proportional to the number of switched-off kernels.

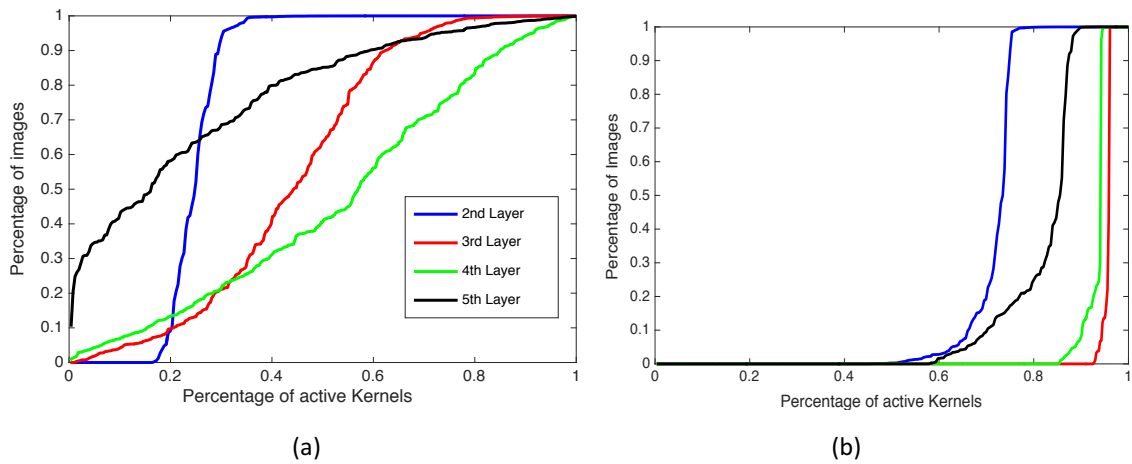

Figure 6: Kernel Activity Profile for CaffeNet: (a) ILSVRC 2012 dataset, (b) FOOD 101 dataset. A large part of kernels remains permanently inactive.

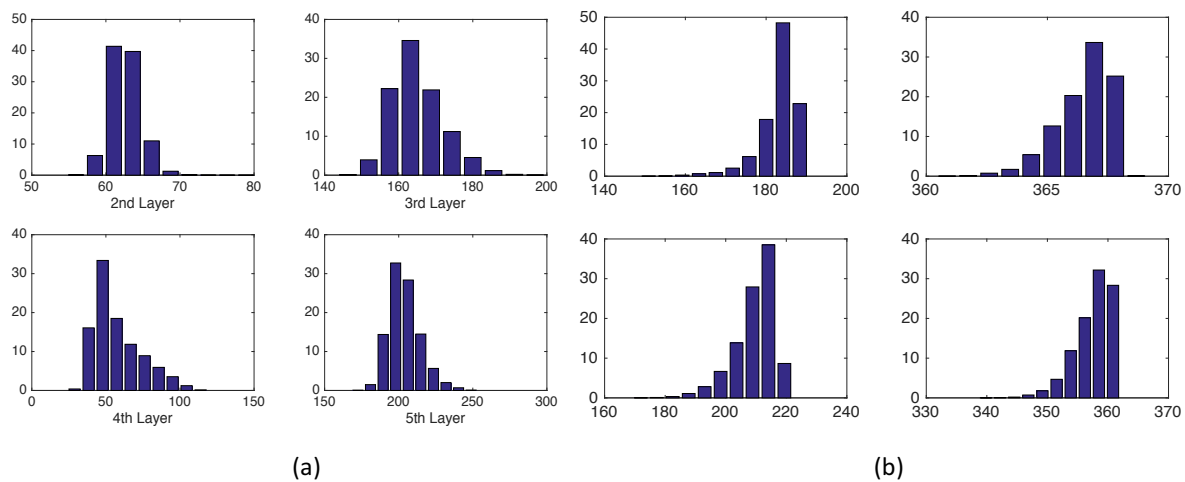

Figure 7: Percentage distribution of the number of inactive kernels throughout the validation set for every layer of CaffeNet, from layer 2 (leftmost) to layer 5 (rightmost): (a) For the ILSVRC 2012 dataset, (b) For the FOOD 101 dataset. The probability is much higher at higher population numbers for the Food 101 dataset (right-skewed), and this is a strong indication of excess redundancy suppression.

A statistical analysis on the switching activity of the CaffeNet for ILSVRC 2012 dataset reveals that only 64.27% of the network's kernels are active (on average) throughout the validation set (35,73% reduction). Specifically, 75.55% of the layer 2 kernels, 57.03% of the layer 3 kernels, 46.93% of the layer 4 kernels and 77.58% of the layer 5 kernels are activated (on average).

The reduction of the computational load In terms of the total MAC operations has been computed to be at 38.31% compared to the reference CaffeNet, taking also into account the computational overhead introduced by the switching modules.

The same analysis on the FOOD 101 dataset indicates that throughout the validation set, only 14.52% of the kernels are activated in the layers of the network. Specifically, 28.49% of the layer 2 kernels, 4.55% of the layer 3 kernels, 6.91% of the layer 4 kernels and 18.15% of the layer 5 kernels are activated on average. The reduction in required MACs in relation with the original CaffeNet model is 72.30%.

***Evaluation on SqueezeNet 1.1,*** has also been conducted on ILSVRC 2012 and FOOD 101 datasets. On the ILSVRC 2012 dataset, the engagement of the LKAM modules results again to the reduction of the kernel filtering operations. This is shown graphically in Figure 8.

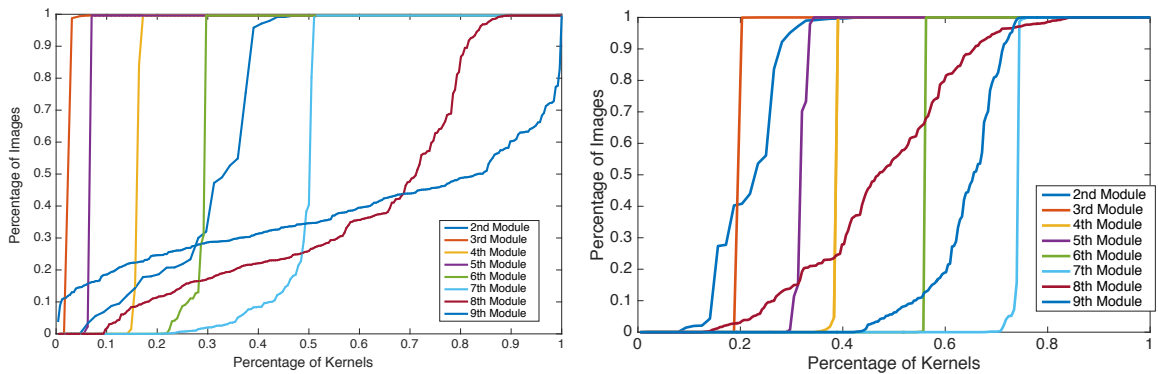

Figure 8: Kernel Activity Profile for every layer (fire module) of SqueezeNet 1.1: (a) ILSVRC 2012 dataset, (b) FOOD 101 dataset. For the easier FOOD 101 problem kernels are mostly either permanently active or inactive.

A statistical analysis on the switching activity of the SqueezeNet 1.1 for the ILSVRC2012 dataset reveals that on average throughout the validation set, only 68.28% of the kernels are active in the layers of the network (31.72% reduction). The reduction in required MACs in relation with the original SqueezeNet model is 30.6%.

A similar analysis on the switching activity of the SqueezeNet 1.1 for the FOOD 101 dataset reveals that on average throughout the validation set, only 56,26% of the kernels are active in the layers of the network (43.74% reduction). The reduction in required MACs in relation with the original SqueezeNet model is 33.4%.

It becomes evident from the results for both CaffeNet and SqueezeNet networks, that the proposed architectural modification results into a significant reduction of the active kernels during inference time. Of equal importance is the fact, that the networks adapt their form and size, depending on the data and the complexity of the classification problem: CaffeNet demonstrates a decrease in the average number of active kernels necessary for carry out the recognition on the FOOD 101 dataset compared to that of the ILSVRC2012 dataset, regardless of being trained with the same regularization gains (eq. 5). That means that the corresponding network features an excess learning capacity, not needed for carrying out this classification task, having got recognized as such and automatically eliminated by the presented training scheme.

***Evaluation on SqueezeNet 1.1, with complex bypass.*** *In order to further push the margins of redundancy reduction,* testing the ability of the proposed method to disable whole convolutional layers, we used the paradigm of SqueezeNet 1.1 with a complex bypass **[Squeeze]**, featuring feed-forward paths. In this structure, each LKAM module controls a 3x3 and a 1x1 filter, so as to make feasible a potential deactivation of every unit within a fire module. To test this configuration, we incorporated a simpler classification problem. To this end, we formulate a day/night detection problem by utilizing the Flickr25K dataset [Flickr] as follows: For the class `night`, we used the special category night of that dataset. For the class `day`, we use images that do not belong to the night class and do not also classify as `water`, `structures`, `indoor` and `transport`. The new dataset contains roughly 5K images and has been balanced to contain equal number of day and night examples.

On this dataset, engagement of the LKAM modules results on a reduction of the kernel filtering operations. This is shown graphically in the Figure 9.

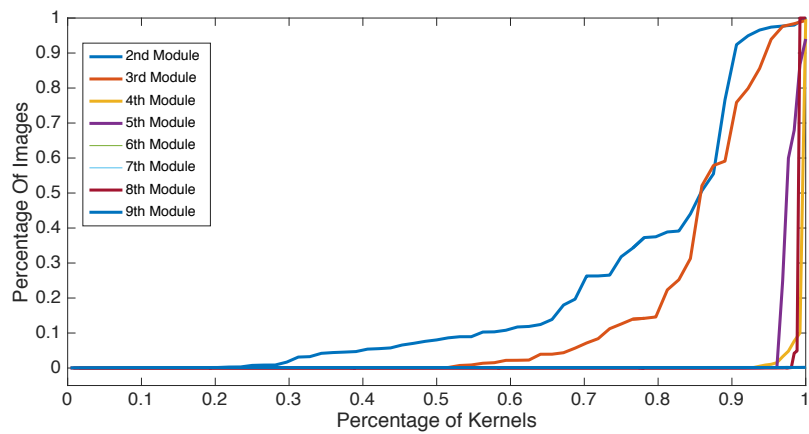

Figure 9: Kernel Activity Profile for every layer (fire module) of SqueezeNet 1.1 with complex bypass, on the Day/Night subset based on the Flickr25K dataset.

From Figure 9 it is discernible that the new structure responds well to the simplification of the classification problem. A statistical analysis throughout the validation set, reveals that on average, only 5.34% of the kernels of the structure are active while the recognition rate is as high as 90.5%. In addition, two fire modules ($7^{th}$ module and $8^{th}$ module) are never used and so they can be permanently pruned. This is an additional indication of the power of the proposed approach for further simplification.

Another interesting fact is that a statistical analysis of the kernel activations of CaffeNet throughout the ILSVRC1012 dataset reveals that images of similar categories tend to use similar kernels. This trend becomes more evident towards the deeper layers, where kernels capture higher-level and more specialized information. Those facts imply that data featuring similar visual attributes, use sub-networks explicitly formed inside the overall trained architecture via the learned activation rules. This enables the forward flow of information in a way which is tailor-made to each sub-class of visual attributes, enabling a parsimonious yet accurate inference process. Furthermore, those findings indicate that LKAMs of a layer could benefit from receiving the SW signals from previous layer as inputs, acting as priors to the kernel activations of the current layer. This path is currently open for future investigation.

## 5. Inference speed Measurements on modern mobile systems

The validity of the proposed approach has been verified for the CaffeNet architecture on three different platforms: Intrinsyc MDP 820, Xiaomi Redmi Note 4 and Samsung S7 Edge. For each implementation, the inference time has been measured and the results are shown in Table 5.

In all these platforms the computations have been offloaded to the GPU while CPU is mostly used for housekeeping functions. GPU is programmed in OpenCL using hand-optimizations aiming to avoid any pre- and post- processing operations at convolutions layers, minimize memory usage by avoiding temporary memories, reduce as much as possible the data transfers from/to GPU and efficiently exploit the de-activation of the kernels.

As it becomes apparent from this table, there is always an important speedup due to the economy in the computations of filtering kernels. However, the speed-up although proportional to the reduction of the total MACs, is not equal to that. This is because of the computational overhead related to the real-time restructuring of the data and kernel tensors within the GPU and prior to the computations, which is necessary for accommodating kernel switching capabilities.

The overhead for the implemented approaches has been analyzed and the results are depicted in Figure 9.This figure illustrates the total inference processing time against the percentage of the total active kernels, which in this case are manually activated to the required percentage.

Table 5: inference speed of CaffeNet on ILSVRC 2012 and FOOD 101 datasets (Batch 12)

| Image Dataset | Mean, **baseline** inference time | Mean, **accelerated** inference time | **Speedup** of the **convolutional part** (times) | **Speedup** of the **overall** inference time (times) |
|---|---|---|---|---|
| Intrinsyc MDP 820 Tablet: Qualcomm SnapDragon 820 (Kryo ARM/Adreno 530) | | | | |
| ILSVRC 2012 | 54.9 ms | 50.7 ms | x1.14 | x1.08 |
| FOOD 101 | 55.5 ms | 30.2 ms | x3.48 | x1.84 |
| Xiaomi Redmi Note 4: Mediatek MT6797 Helio X20 (ARM/Mali-T880 MP4) | | | | |
| ILSVRC 2012 | 250.7 ms | 190.0 ms | x1.36 | x1.32 |
| FOOD 101 | 250.2 ms | 107.1 ms | x2.64 | x2.34 |
| Samsung S7 Edge Exynos 8890 Octa (ARM/MALI-T880 MP 12) | | | | |
| ILSVRC 2012 | 110 ms | 78.9 ms | x1.53 | x1.39 |
| FOOD 101 | 110 ms | 36.4 ms | x5.49 | x3.02 |

From this figure, the following conclusions can be derived:

- There is an overhead imposed by the computation load, which is related with the manipulation of data, and the computations associated with the functionality of the LKAM modules.
- This overhead, expressed both as processing time and as percentage of the nominal computational time, is depended on the computational platform.
- The processing time is linearly proportional to the number of active kernels.
- This proportionality depends on the target platform.
- Regardless of the imposed overhead, there is a clear advantage in following a dynamic inference approach, in the context of the presented scheme.

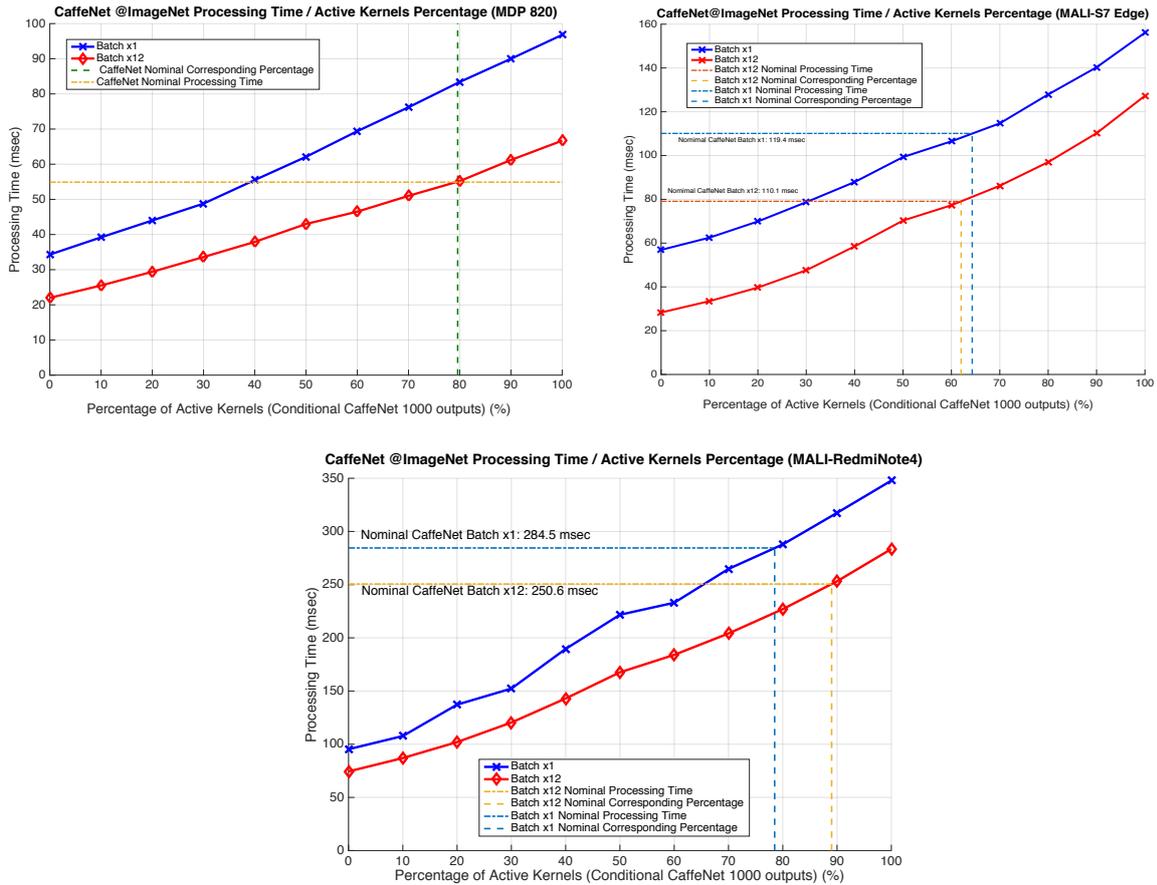

Figure 9: Processing time vs the percentage of the total active kernels for CaffeNet on the: ILSVRC 2012 dataset. The nominal performance (e.g. the processing time of the original network) is also depicted in each case.

## 6. VLSI hardware Implementation

The proposed method brings significant advantages in the case of a VLSI hardware implementation, where all the filter kernels are to be implemented as separated hardware blocks in a parallel architecture where all the filter kernels are operating on the same feature map.

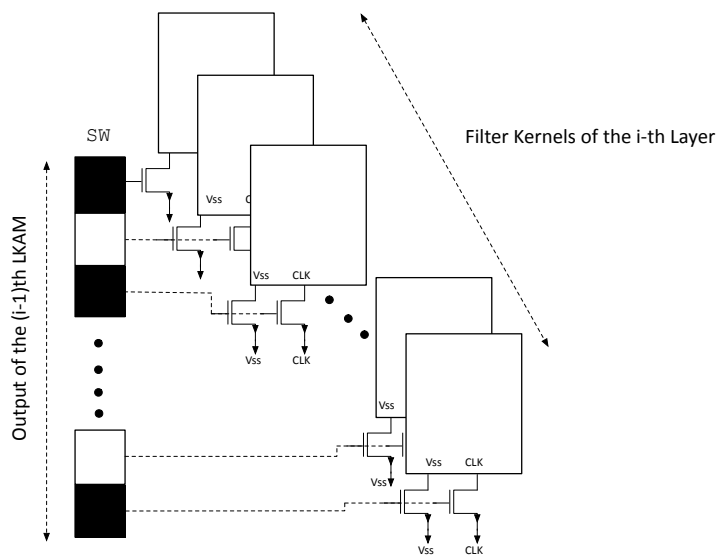

Figure 10: Example of use of the LKAM switch in VLSI implementations.

In this case, the LKAM can be implemented as an array of switches by the virtue of a set of voltage controllable switches (i.e. CMOS transistors) as shown in Figure 10. In this case, the circuitry implementing a filter kernel can be switched-off, by cutting feed in both power and clock inputs, saving this way the energy corresponding to both its dynamic and static bias consumption.

**Conclusions**

We presented a systematic way for implementing CNN variants that are parsimonious in computations. A new CNN designing approach has been proposed which allows a CNN to learn to use as few as possible computing resources and change size and form during inference in real-time, depending on the input data.

The proposed framework incorporates a new learning module, the Learning Kernel Activation Module (LKAM), able to occasionally or permanently de-activate a sub-set of filtering kernels depending on the input image content during the inference phase. Using this new module, the CNN learns during the training phase how to reduce its size in real time and thus result in a significant computational economy.

Simulation results (using Caffe framework) and real-time measurements of optimal embedded implementations, verify the efficacy of the proposed method and demonstrate the ability of the resulting networks to adapt their size to the complexity of the classification task and the sparsity of the input data.